%% file: sparsedeep.tex
\documentclass[10pt,twocolumn,letterpaper]{article}

\pdfoutput=1

\usepackage{fakecvpr}
\usepackage{times}
\usepackage{epsfig}
\usepackage{algorithmic}
\usepackage{multirow}

\usepackage{mydefs}

\usepackage{rotating}

\usepackage[pagebackref=true,breaklinks=true,letterpaper=true,colorlinks,bookmarks=false]{hyperref}

\begin{document}

\title{Memory Bounded Deep Convolutional Networks}

\author{Maxwell D. Collins$^{1,2}$ \\ $^1$ University of Wisconsin-Madison \\ \texttt{mcollins@cs.wisc.edu}
  \and Pushmeet Kohli$^{2}$ \\ $^2$ Microsoft Research \\ \texttt{pkohli@microsoft.com}}
\date{}
\maketitle

\input{abstract.tex}

\input{intro.tex}
\input{related-work.tex}

\input{cnns.tex}
\input{updates.tex}

\input{fig/kernels/kernels.tex}
\input{implementation.tex}

\input{experiments.tex}

\input{discussion.tex}


{\small
\bibliographystyle{plain}
\bibliography{sparsedeep}
}

\appendix
\input{supp-memory.tex}

\end{document}

%% file: abstract.tex
\begin{abstract}
  In this work, we investigate the use of sparsity-inducing regularizers during training 
  of Convolution Neural Networks (CNNs). These regularizers encourage that fewer connections in the convolution and 
  fully connected layers take non-zero values and in effect result in sparse connectivity 
  between hidden units in the deep network. This in turn reduces the memory and runtime 
  cost involved in deploying the learned CNNs. We show that training with such regularization
  can still be performed using stochastic gradient descent implying that it can be used 
  easily in existing codebases. Experimental evaluation of our approach on MNIST, CIFAR, and ImageNet 
  datasets shows that our regularizers can result in  dramatic reductions in memory requirements.
  For instance, when applied on AlexNet, our method can reduce the memory consumption by a factor of four with minimal loss in accuracy.
\end{abstract}
%

%% file: intro.tex
\section{Introduction}
Over the last few years, high capacity models such as deep Convolutional Neural
Networks (CNNs) have been used to produce state-of-the-art results for a wide
variety of vision problems including Image classification and Object detection.
This success has been in part attributed to the feasibility of training models
with large number of parameters and the availability of large training datasets.

Recent work has made significant strides in the techniques used to train deep
learning models, making it possible to optimize objective functions defined over 
million of parameters. These techniques however require careful tuning of the optimization
and initialization hyperparameters to ensure that the training procedure arrives
at a reasonable model \cite{krizhevsky2012-imagenet-cnn}.
In addition, simply performing the computations required for models with
so many parameters has required leveraging high-performance parallel
architectures such as GPUs \cite{jia2014-caffe,krizhevsky2012-imagenet-cnn},
or distributed clusters \cite{dean2012-distbelief}.

While the capacity of such deep models allows them to learn sophisticated mappings,
it also introduces the need for good regularization techniques.
Furthermore, they suffer from high memory cost at deployment time due to large model sizes.
The current generation of deep models \cite{krizhevsky2012-imagenet-cnn,lin2013-network-in-network, szegedy2014-googlenet}
show a reasonable degree of generalization in part because of recent advances in 
regularization techniques like DropOut~\cite{krizhevsky2012-imagenet-cnn}. However,
these and other approaches~\cite{krogh1991-weight-decay} proposed in the literature fail to address the issue of high memory cost, which is particularly 
problematic in models that rely on multiple fully connected layers.
The high memory cost becomes especially important when deploying in
application in
computationally constrained architectures like mobile devices.

\input{fig/memory/memory-combined.tex}

To overcome the aforementioned problems, we propose the use of sparsity-inducing regularizers
for Convolution Neural Networks (CNNs). These regularizers encourage that fewer connections
in the convolution and fully connected layers take non-zero values and in effect result 
in sparse connectivity between hidden units in the deep network. In doing so, the regularizers
not only restrict the model capacity but also reduce the memory and runtime
cost involved in deploying the learned CNNs.

We applied our method on MNIST, CIFAR and ImageNet datasets.
Our results show that one
can generate models that reduce the memory consumption by a factor of 3 or more
with minimal loss in accuracy.
We show how this can be used to improve the accuracy of vision classifiers
using ensembles of deep networks \emph{without} incurring the greater
memory costs that would ordinarily result from having to store multiple
models.
Using sparsity regularization in this way significantly improves
upon more typical ways researchers seek to limit model complexity,
by e.g.~changing the network topology by removing units or neurons.
Finally, we show how the regularized training objectives can be efficiently
optimized using stochastic gradient descent.

The key \textbf{contributions} of this paper are:
\begin{enumerate}
  \item A set of sparsity-inducing regularization functions that we demonstrate are effective at reducing model complexity with no or minimal reduction in accuracy.
  \item Updates for these regularizations that are easily implemented within standard existing stochastic-gradient-based deep network training algorithms.
  \item Empirical validation of the effect of sparsity on CNNs on CIFAR and MNIST datasets.
\end{enumerate}


%% file: fig/memory/memory-combined.tex
\begin{figure}
  \begin{center}
    \begin{tabular}{c c}
      \hspace{0.08\linewidth} MNIST & \hspace{0.09\linewidth} CIFAR-10 \\
      \includegraphics[width=0.45\linewidth, clip=true, trim=0 0 0 0]{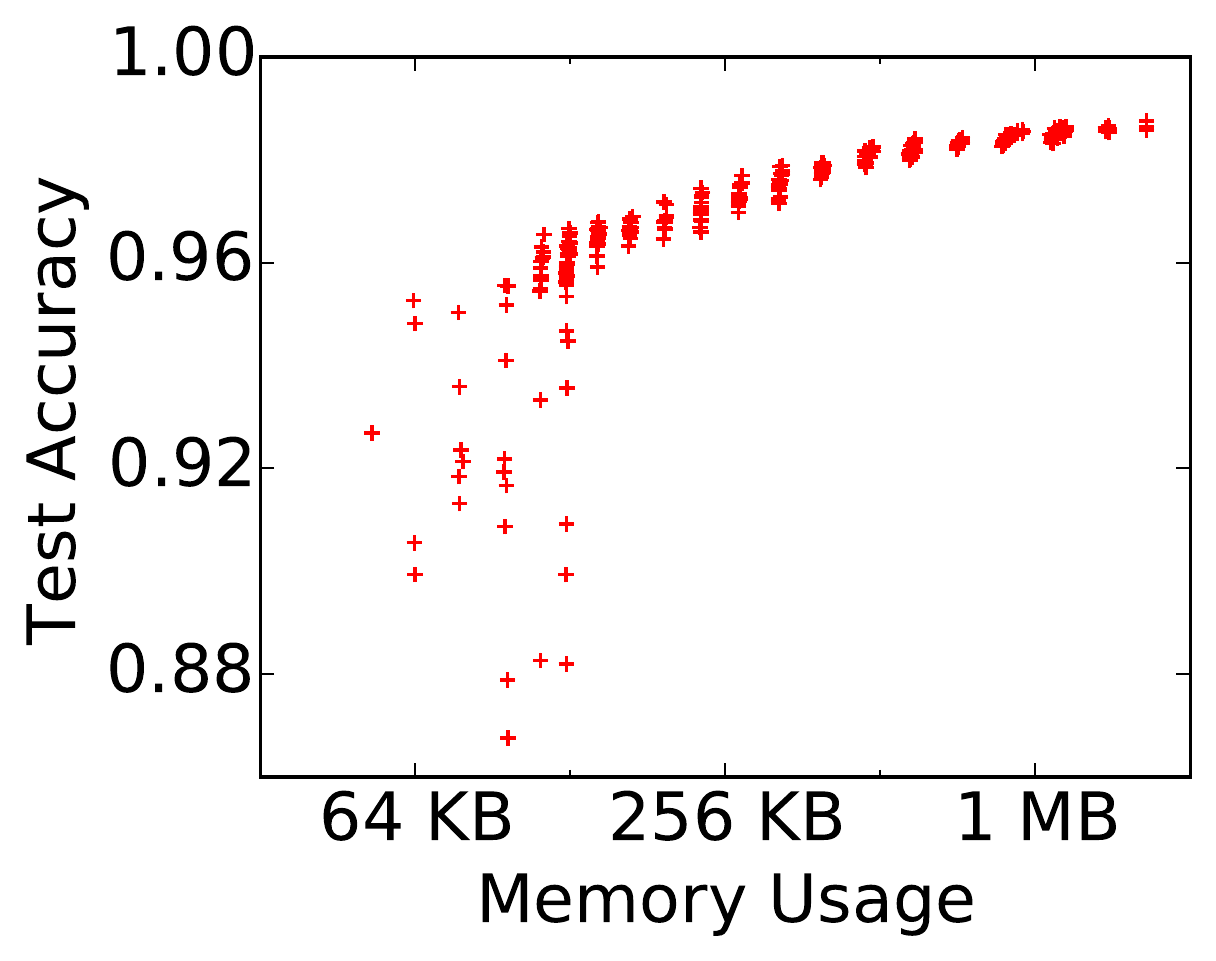}
      & \includegraphics[width=0.45\linewidth, clip=true, trim=0 0 0 0]{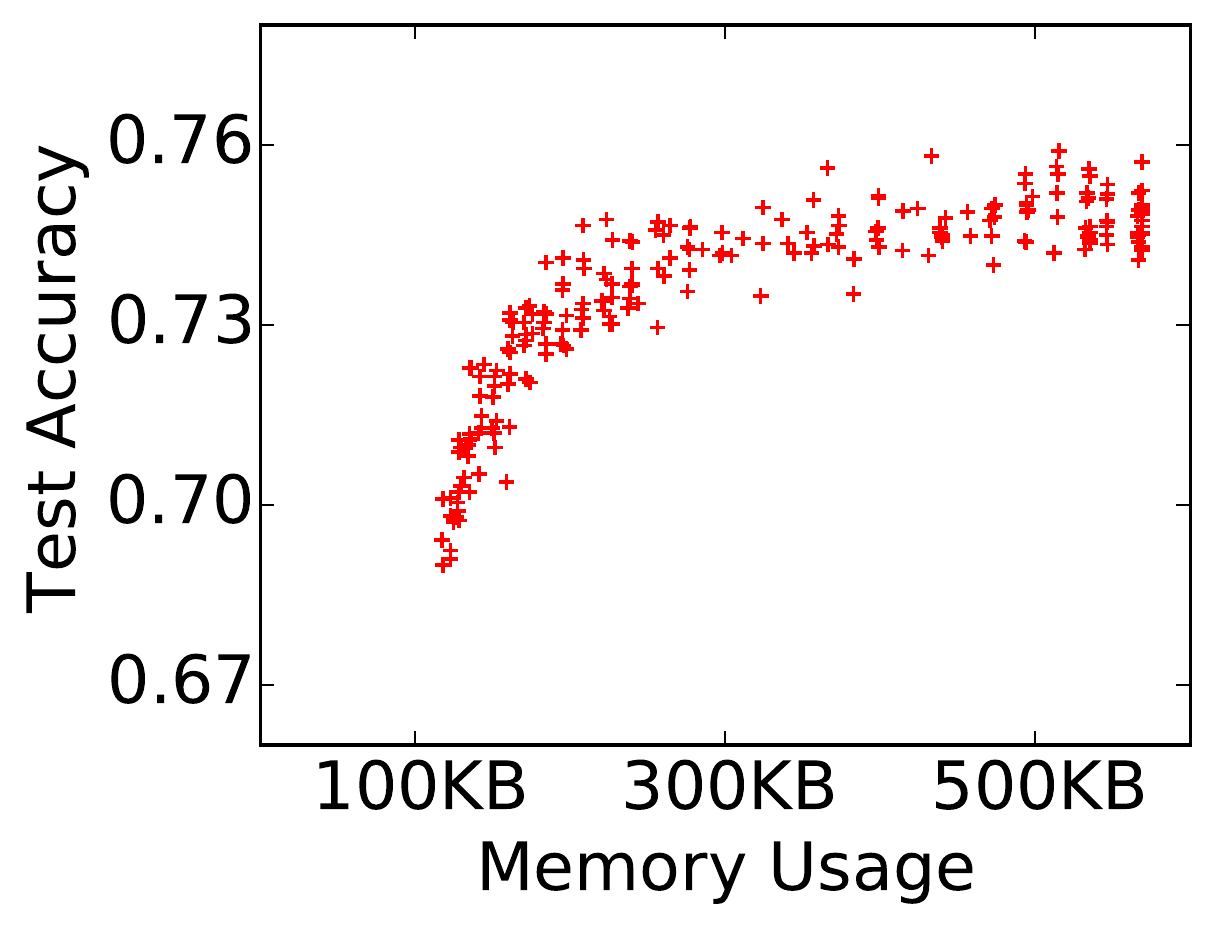}
    \end{tabular} \vspace{0.5em} \\
    ImageNet \\ \vspace{0.5em}
    \begin{tabular}{l l l l}
      \textbf{Model} & \textbf{Top-1} & \textbf{Top-5} & \textbf{Memory} \\
      \hline
      Reported \cite{krizhevsky2012-imagenet-cnn} & 59.3\% & 81.8\% & - \\
      Caffe Version \cite{jia2014-caffe} & 57.28\% & 80.44\% & 233 MB \\
      Sparse (ours) & 55.60\% & 80.40\% & 58 MB \\
    \end{tabular}
  \end{center}

  \caption{ \label{fig:memory}
    Exploring the trade-off between accuracy and model size achieved by our
    method for different problems.
    In the first row, each point plotted is a candidate network
    considered by the greedy search in Section \ref{sec:greedy-layerwise}.
    The $x$ axis shows the memory required to store the weights of that
    network, using the best choice of the storage formats described
    in Appendix \ref{sec:memory}.
    In the table we show the same for a network trained on ILSVRC 2012,
    as described in Section \ref{sec:implementation}.
  }
\end{figure}

%% file: related-work.tex
\subsection{Related Work}

\paragraph{Regularization of Neural Networks}
Weight decay was one of the first techniques for regularization of neural
networks and was shown to significantly improve the generalization
of Neural Networks by Krogh and Hertz \cite{krogh1991-weight-decay}
It continues to be a key component of the training for
state-of-the-art deep learning methods \cite{krizhevsky2012-imagenet-cnn}.

Weight decay works by reducing the magnitude of the weight vector of the network. 
It yields a simple additional term to the weight updates at each iteration
of the learning procedure. The update term used is the gradient of the squared $\ell_2$-norm
of the weights. An interesting observation is that the training procedure for a
linear perceptron with weight decay is equivalent to learning a linear SVM using stochastic gradient descent~\cite{shalev-shwartz2011-pegasos}, where the weight decay is serving as the update to maximize the margin of the classifier.

While not seen in CNNs or Computer Vision application,
building sparse networks was considered for more classical neural networks.
These were grouped into ``pruning'' methods \cite{reed1993-pruning},
and include both regularization penalties and techniques for determining
the importance of a given weight to the network's accuracy.

Hinton {\em et. al}~\cite{hinton2012-dropout} proposed a regularization
technique known as \emph{dropout}, which now forms a basis for many 
state-of-the-art deep neural network models~\cite{krizhevsky2012-imagenet-cnn,lin2013-network-in-network,szegedy2014-googlenet}
During training, some portion of the units in the network are ``dropped:''
their output is fixed to zero and their weights are not updated during
back-propagation. The output of these units are then multiplied by the dropout factor at test
time. This is done in part as a computationally cheaper approximation to training an
ensemble of sparse networks.

The sparse networks implicitly created by dropout are simplified in a direction
orthogonal to the parameter regularization considered here.
Dropout eliminated entire units or neurons, while we seek to reduce the number of parameters of those units.
Rather than having a smaller number of complex units,
we try to train models that have a constant number of simpler units.

\paragraph{Comparison to Model Compression}
A further method of constructing simpler models is the technique of
\emph{model compression} \cite{bucilua2006-model-compression}.
Model compression relies on the availability of large amounts of unlabelled
data, using it to build smaller and computationally cheaper models by
teaching the compressed model from the output of a larger and more
parameter-heavy model that has been trained to fit the target task.
This data is labelled using the original network and then used to train a
smaller network.
The idea behind this is to make sure that the smaller network does not have to
worry about regularization.
Furthermore, unlike our method where sparsity is enforced explicity through the
use of an $\ell_0$ or $\ell_1$ norm on a large number of hidden units,
the model compression work simply trains a network with smaller number of
hidden units.
Like Dropout regularization,
it essentially maintains the dense connectivity structure
between layers while our method results in sparser connectivity between layers.
In this work we target a similar setting,
in which plenty of computational resources
are available at training time, but we wish to use these to train a model
that is suitable for deployment in a more constrained environment.

Some recent work \cite{jaderberg2014-low-rank-cnn,zhang2014-low-rank-cnn}
has considered the same task but have adopted a different approach.
They build networks with a set of \emph{low rank} filters in each layer.
This is built after training a network without this constraint,
where the simpler network is selected to approximate the original full-rank
network.

\paragraph{Sparse Autoencoders}
Extensive work has also been done on training deep models that learn
sparse \emph{representations} of the data, while the learned parameters are themselves nonsparse.
In an early work on models with this prior, Olshausen and Field~ \cite{olshausen1997-sparse-coding} used a basic
neural network structure inspired by the V1 layer of the visual cortex
to learn sparse representations of a set of images.
More recent work has used the term \emph{sparse autoencoder} to describe
this type of structure.
and in deep learning autoencoders have been used to initialize multi-layer
networks as a method of layer-wise pre-training \cite{vincent2008-denoising-autoencoders}.

Indeed, much recent work on designing novel architectures for deep learning
for vision application seeks to reduce or sparsify the parameter set.
Convolutional layers themselves, in addition to enforcing an assumption of
translation invariance,
are meant in part to simply reduce the number of parameters \cite{lecun1998-document-recognition}.
A convolution \emph{can} be represented by a fully-connected layer,
though it is unlikely to learn this mapping with a finite amount of training
data and computation time.
Some prior work has investigated automatically determining CNN structure,
though instead by searching the space of layers and their sizes
that may be used in place of simpler higher-parameter-count choices \cite{snoek2012-bayesian-optimization}.

%% file: cnns.tex
\section{Convolutional Neural Nets}

Convolutional Neural Nets are a variety of Deep Learning methods in which
\emph{convolutions} form the early layers.
These have become the standard technique for performing deep learning for
Computer Vision problems,
as they explicitly deal with vision-based primitives.
In the first of these layers learned filters are convolved with the input image.
The output of each of these kernels, usually composed with some nonlinear
mapping, is then convolved with another set of
learned filters in the second layer.
Each subsequent convolution layer then convolves another set of filters
with the output of each convolution in the previous layer.

CNNs are frequently augmented with final ``fully connected'' layers
that are more similar to classical neural networks.
These provide a mapping from the features learned by the convolutional
layers to the output labels.
In most state-of-the-art networks the fully-connected layers are responsible
for the majority of the parameter count of the network,
though some recent work has considered models consisting entirely of
convolutional layers with a novel architecture
\cite{lin2013-network-in-network}
or converted the fully-connected layers into a convolution \cite{sermanet2014-overfeat}.
The test models we consider for image classification
will have in their final layer a number of units equal to the number of
output classes for the benchmark task.
Each unit will correspond to an output class,
and the unit that produces largest output on a given input image will
be the class label the model predicts for that image.

In this work, the terms ``weights'' and ``parameters'' are
used interchangeably to refer to
the trained weighting on the signal sent from one unit to another connected
unit in the next layer in forward propagation.
The set of parameters of a CNN also includes the individual elements per
pixel and channel in a learned convolution filter.
In a classic ``fully connected'' or ``inner product'' layer, each unit
in the layer will take the inner product of its weight vector and the output
of all units in the previous layer.
This may then be transformed by some nonlinear function, or simply output
as is to the next layer.
Individual CNN units will frequently also have an additive ``bias,''
which we also count in our parameter costs but contribute little to the overall
size and complexity of most networks.

\subsection{Optimization Perspective on CNN Training}

The standard methods for training deep learning models may be expressed as
stochastic gradient descent on a given loss function between the output
prediction and training labels.
The training procedure is given a set of input vectors
$\mathbf{x}_1, \mathbf{x}_2, ..., \mathbf{x}_n$
and the corresponding ground truth labels $y_1, y_2, ..., y_n$.
A basic feed-forward network can be treated as a function $f$ of the input data
and the collection $W$ of learned parameters for all layers.
The overall optimization problem on these weights may then be posed as follows:
\begin{equation}
  \min_W \; \left( \mathcal{O}(W) := \frac{1}{n} \sum_{i=1}^n \mathcal{L}(y_i, f(\mathbf{x}_i, W)) + \lambda r(W) \right)
  \label{eq:dnn-opt}
\end{equation}
Here, $\mathcal{L}$ is a loss function between the true labels $y_i$
and the predictions of the network $f(\mathbf{x}_i, W)$ on the
$i^\text{th}$ training example.
Taking the sum over the training data is treated as a proxy for the empirical
risk.
The function $r$ is the \emph{regularization} term,
with a weighting hyperparameter $\lambda$,
which seeks to reduce the hypothesis space.
For ordinary weight decay, this will be a squared $\ell_2$ norm:
$r(W) = \|W\|_2^2$.
For image classification the loss $\mathcal{L}$ will be a soft-max loss.

The objective in \eqref{eq:dnn-opt}, along with its gradients,
is very expensive to compute for practical problems.
The state of the art in Computer Vision considers very large networks
with up to 22 layers \cite{szegedy2014-googlenet}
and 60 million parameters \cite{krizhevsky2012-imagenet-cnn}.
Further, these models are trained on large datasets such as ImageNet,
with the 2012-2014 classification challenge set having 1.2 million training images.
The optimization of this objective is made far more computationally tractable
by using \emph{stochastic gradient descent} \cite{bottou2010-large-scale-ml-sgd}
or stochastic variants of
adaptive and accelerated gradient methods \cite{sutskever2013-momentum}.

Instead of using the full objective $\mathcal{O}$,
these approximate the gradient using a sample $B \subset \{1,...,n\}$
drawn from the training data, taking instead the gradient of the function:
\begin{equation}
  \mathcal{O}_B(W) := \frac{1}{|B|}
  \sum_{i \in B} \mathcal{L}(y_i, f(\mathbf{x}_i, W)) + \lambda r(W)
  \label{eq:stoch-obj}
\end{equation}
This \emph{stochastic} gradient has the key property that it is in expectation
equal to the true gradient of the full objective function:
\begin{equation}
  \mathbb{E}\left[ \grad \mathcal{O}_B(W) \right]
  = \grad \mathcal{O}(W)
\end{equation}

%% file: updates.tex
\section{Regularization Updates}

We encourage sparsity in the networks by applying simple updates to
the set of weights in each layer during training.
First, though, we consider the regularization functions themselves.
Regularization is typically based on a norm function on the parameters
of the model taken as a vector.
Define the $\ell_p$ norms as
\begin{equation}
  \|x\|_p = \left( \sum_i |x_i|^p \right)^{1/p}
\end{equation}
for $0 < p < \infty$.
This definition is typically extended to $\ell_0$ and $\ell_\infty$ norms
by taking limits on $p$.
Of special interest to sparsity regularization is the $\ell_0$ ``norm,''
as $\|x\|_0 = \lim_{p \to 0} \|x\|_p$ is the count of the number of nonzeros
of $x$.


\subsection{$\ell_1$ Regularization and the Shrinkage Operator}

The most common classical technique for learning sparse models in
machine learning is $\ell_1$ regularization \cite{tibshirani1996-lasso}.
The $\ell_1$ norm of a vector is the tightest convex relaxation of the
$\ell_0$ norm.
It has been shown for some classes of machine learning models
that regularization terms consisting of an $\ell_1$ norm can provide a
provably tight approximation, or even an exact solution,
to a corresponding $\ell_0$ regularization that directly penalizes or
constrains the number of nonzero parameters of the model
\cite{donoho2006-compressed-sensing}.

We can optimize an $\ell_1$ regularization term by updating the weights
along a \emph{subgradient} of the $\ell_1$ norm.
A negative multiple of a subgradient gives a descent direction,
and updating the weights a sufficiently small distance along this ray
will reduce the regularization term.
Its sum with the gradient of the loss terms in \eqref{eq:dnn-opt}
gives a subgradient of the whole objective.
Since the $\ell_1$ norm is differentiable almost everywhere,
so the subdifferential is a singleton set,
it is not even usually necessary in practice in CNN training
to choose a particular subgradient.
Therefore, the choice of subgradient we investigate in the paper is simply
the sign operator, applied element-wise to the vector of weights.
It can be seen that
$\mathrm{sign}(W) \in \partial r(W)$ for $r(W) = \|W\|_1$.
The resulting update is therefore, for some $\delta > 0$,
the element-wise operator:
\begin{equation}
  W_i \gets W_i - \delta \mathrm{sign}(W_i)
\end{equation}
This $\ell_1$ update is currently implemented in Caffe \cite{jia2014-caffe}
as a type of weight decay that can be applied the whole network.

This update has the significant shortcoming that,
while it will produce a large number of weights that are very \emph{near} zero,
it will almost never output weights that are exactly zero.
Since later layers can still receive and learn to magnify input based on
the resulting small nonzero weights,
and earlier layers will receive backpropagated gradients along these weights,
these near-zero parameters cannot be ignored and the model is not
necessarily any simpler.
When seeking to construct a sparse model, the natural technique is to try
to threshold away these very small weights that an optimal $\ell_1$ solution
will have set to zero.
Finding a good threshold for this heuristic requires some attention to the
schedule of learning weights and the level of regularization relative to the
gradients near the end of training.

A regularization step with significantly better empirical and theoretical
properties is the shrinkage operator \cite{tibshirani1996-lasso}:
\begin{equation}
  W_i \gets (|W_i| - \delta)_+ \mathrm{sign}(W_i)
  \label{eq:l1-shrinkage}
\end{equation}
Here the notation $x_+$ refers to the positive component of a scalar $x$.
The shrinkage operator is among the oldest techniques for sparsity-inducing
regularization.
A key property for deep network training is that this operator will not allow
weights to change sign and ``overshoot'' zero in an update.
The operator will output actual zero weights rather than small weights
oscillating in sign at each iteration.
For a suitable choice of $\delta$, it eliminates the need to consider
thresholding.
Neighboring layers will get actually zero input/backpropagation along that
connection, so they can properly learn on the zero weight.

This shrinkage update is an example of a \emph{proximal mapping}
(also called a ``proximal operator'') and alternating this with descent
steps along the gradient of the loss would yield a \emph{proximal gradient}
method.
These methods generalize and extend the LASSO, and have been applied
to e.g. group sparsity regularizations \cite{wright2009-group-sparse-proximal}.
Proximal mappings are also effective when using approximate or stochastic
gradients.
This was demonstrated by Tsuruoka et. al. for $\ell_1$-regularized regression \cite{tsuruoka2009-l1-sgd},
and Schmidt et. al. \cite{schmidt2011-inexact-proximal-gradient}
provide convergence guarantees in the case of a convex problem for both
stochastic gradient and accelerated variants.

\subsection{Projection to $\ell_0$ balls}

While $\ell_1$ regularization is known to induce or encourage sparsity
in shallow machine learning models, the direct way to construct sparse models
is to consider the $\ell_0$ norm.

We propose a simple regularization operator to train models under
$\ell_0$ norm constraints.
Under this update, every $n$ updates we
set to zero all but the $t$ largest-magnitude elements of the parameter vector.
This imposes the hard constraint that $\|W\|_0 \le t$ for some integer $t$.
This operator can be seen as a projection onto an $\ell_0$ ball, namely:
\begin{equation}
  \mathcal{P}_0(W, t) =
  \left(
  \begin{aligned}
    \arg \min_{W'} \quad & \|W - W'\|_2^2 \\
    \text{s.t.} \quad & \|W'\|_0 \le t
  \end{aligned}
  \right)
  \label{eq:l0-projection}
\end{equation}
This ``projection'' matches the operator of hard-thresholding all but the $t$
least-magnitude elements,
as the quantity the projection seeks to minimize
in \eqref{eq:l0-projection}
is the total squared magnitude of the elements that are zero in the optimal $W'$
but not in $W$.

This $\ell_0$-regularization update is inspired by ``projected gradient''
methods frequently used in convex optimization.
In these methods either the iterate or the gradient along which the next
iterate is generated are projected onto a convex feasible set.
Consider the former case, most similar to our $\ell_0$ update.
If the optimum is not in the interior, this basic procedure will search along
the boundary of the feasible set.
A theoretical analysis of projected gradient algorithms can be gleaned from the more general
analysis on proximal operators.
Projection onto a convex set $P$ is the proximal operator for a extended-valued
function:
\begin{equation}
  r(x) = \begin{cases} 0 & \text{if } x \in P \\ \infty & \text{otherwise} \end{cases}
\end{equation}
Therefore, projection onto a convex feasible set can be seen as a special case
of proximal stochastic gradient methods.
See \cite{recht2013-stochastic-matrix-completion} for discussion specifically
on projection within stochastic gradient methods.

Results from convex optimization provide us with theoretical guarantees of the
efficacy of proximal stochastic gradient methods.
In this case neither the objective nor the feasible set onto which we are
projecting are convex.
For this reason, in shallow models the $\ell_0$ norm is typically eschewed for
being intractable to optimize, and optimization problems incorporating
$\ell_0$-norm terms lack many of the guarantees we see with convex models.
When instead training deep learning models, these guarantees are already
absent.
However, recent advances in the training techniques for deep learning have
made optimizing such models for Computer Vision tasks fairly robust.
We empirically see that directly using the $\ell_0$ norm works,
and stochastic gradient solvers successfully optimize the training error.


%% file: fig/kernels/kernels.tex
\begin{figure*}
  \begin{center}
    \includegraphics[height=0.2857\linewidth]{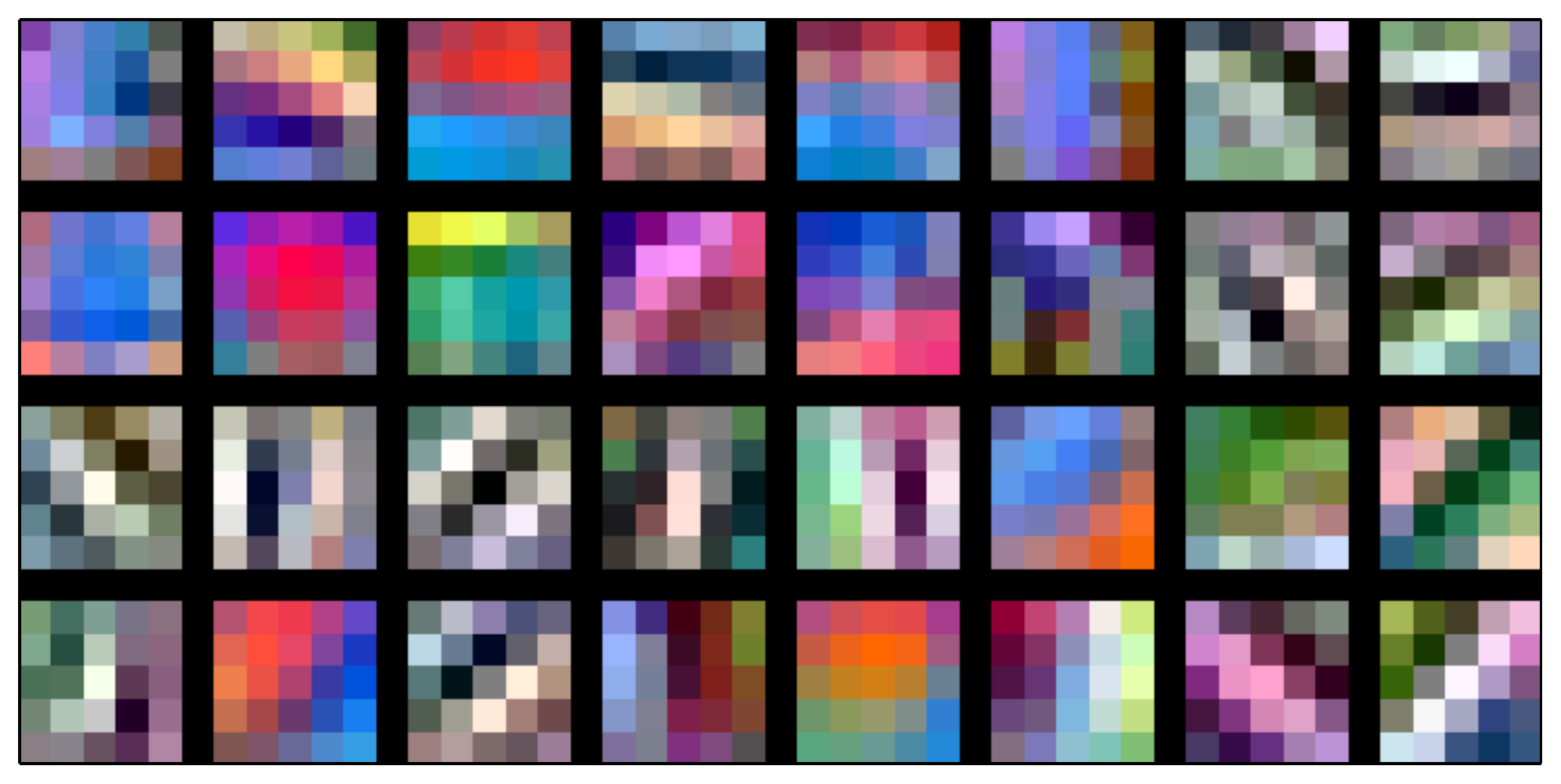}
    \nolinebreak[4]
    \includegraphics[height=0.28\linewidth]{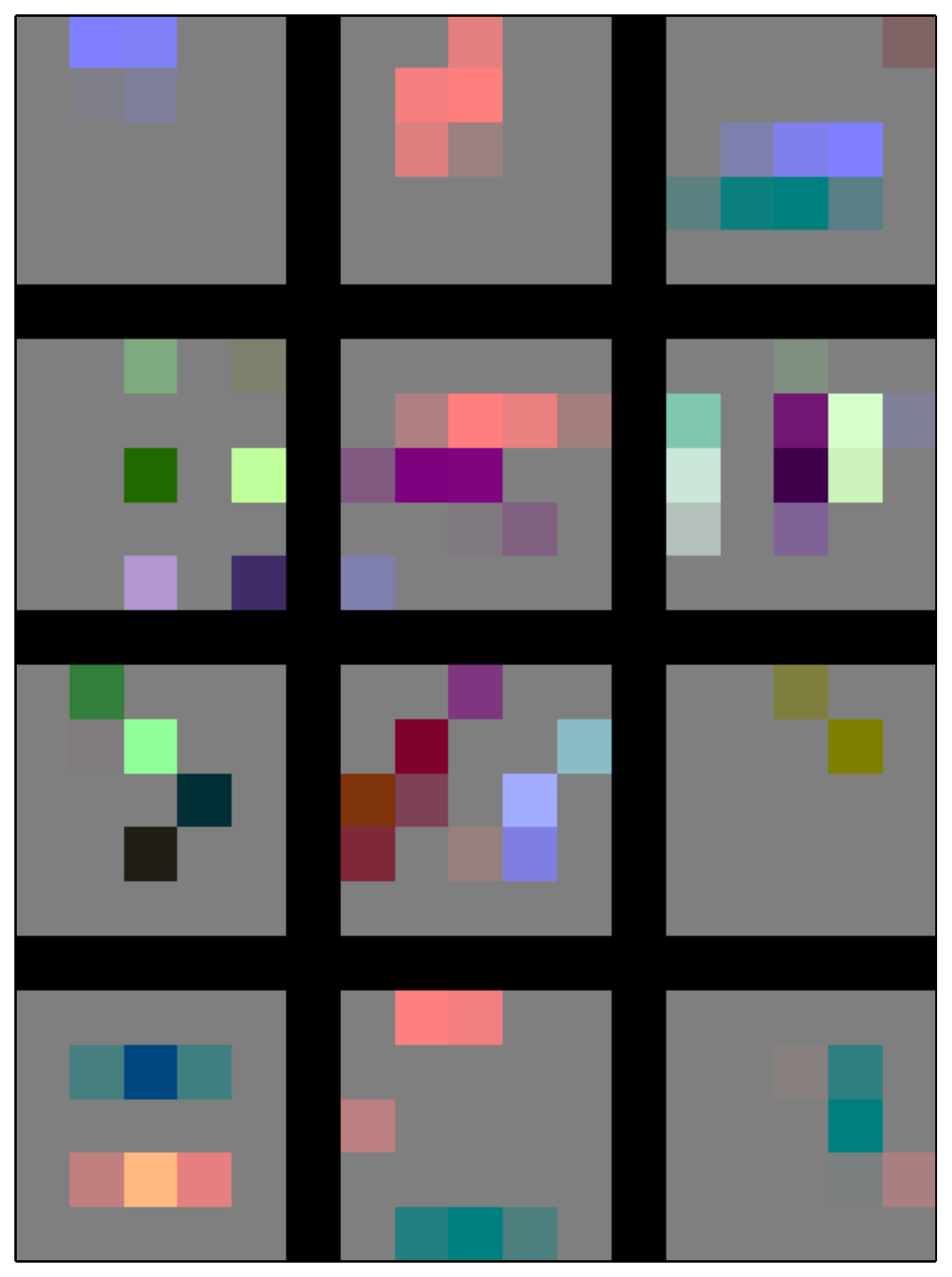}
    \nolinebreak[4]
    \includegraphics[height=0.28\linewidth]{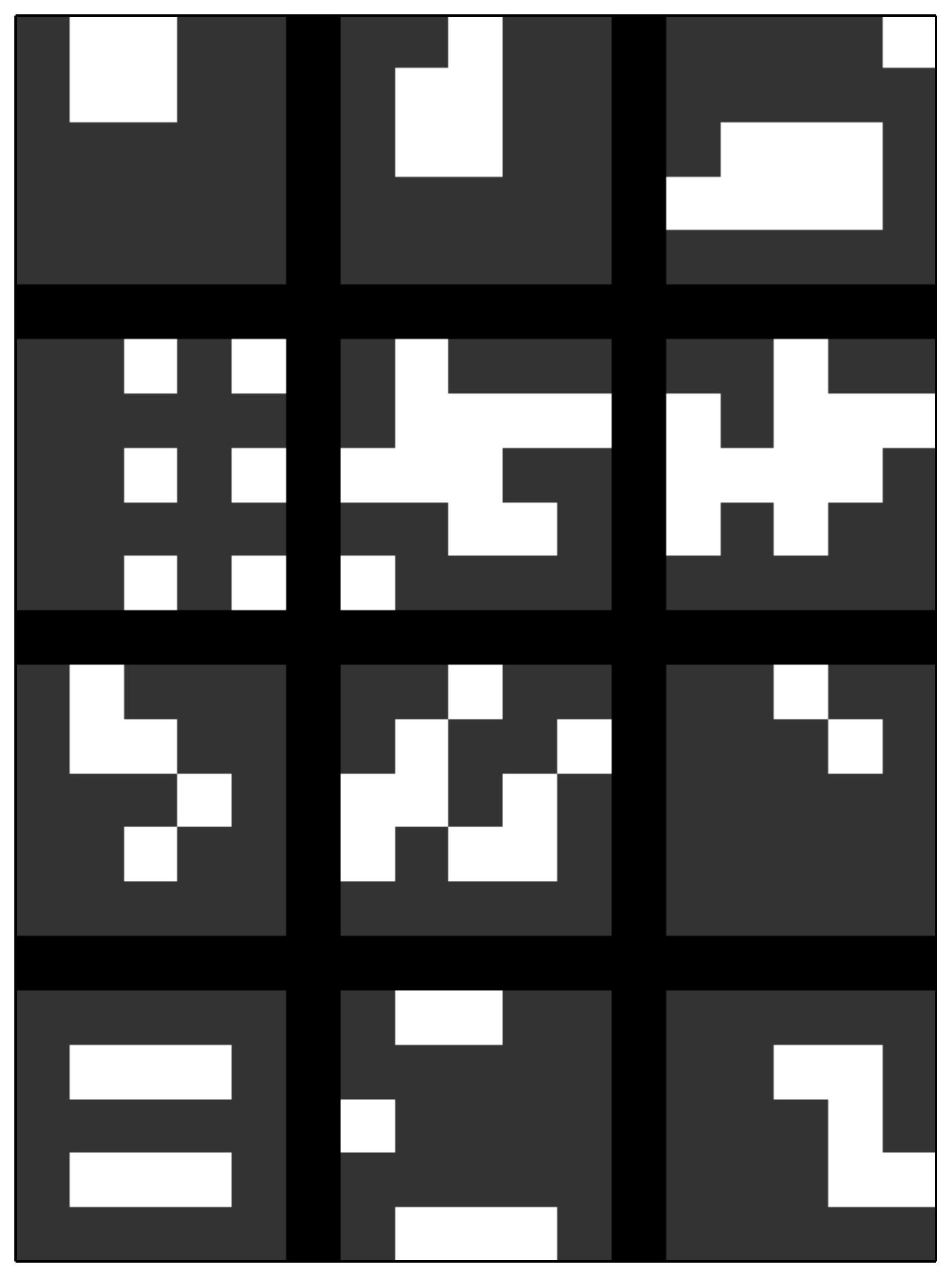}
  \end{center}

  \caption{ \label{fig:kernels}
    Convolution filters learned on the CIFAR-10 classification task.
    The non-sparse kernels on the left come from a baseline model using
    classical weight decay as regularization.
    Incorporating a sparsity-inducing $\ell_1$ shrinkage operator
    during training yields the sparse filters on the middle right,
    and 20 all-zero filters not shown.
    The pixel-wise nonzero pattern of the sparse filters is shown on the
    far right.
  }
\end{figure*}

%% file: implementation.tex
\section{Implementation}
\label{sec:implementation}

Our regularization updates were implemented as a modification to
Caffe \cite{jia2014-caffe}.
The $\ell_1$ update matches that already present, though we extended that
implementation to allow mixing different types of regularization per-layer.
Therefore, for example in Figure \ref{fig:partial-sparsity}
we were able to retain
ordinary weight decay for all but select portions of the model in order to
analyze the effect of sparsity on particular parts of the model.
We use as a target for our regularization the baseline networks provided
with that distribution, LeNet \cite{lecun1998-document-recognition}
and CIFAR-10 Quick \cite{snoek2012-bayesian-optimization}.

For larger-scale experiments, we demonstrate the usefulness of these
regularizers on a network developed for the
ImageNet classification challenge \cite{krizhevsky2012-imagenet-cnn}.
This network has five convolutional layers and three fully-connected layers,
with a total of 60 million parameters.
Upon the publication of \cite{krizhevsky2012-imagenet-cnn},
it was one of the largest neural networks described in the literature.
Due to the much larger scale of both the dataset and the network,
we perform experiments using a computationally cheaper ``fine-tuning''
procedure in which we use an existing non-sparse network as the initialization
for sparsity-regularized training.
The starting point for this training was a Caffe-based \cite{jia2014-caffe}
duplication of the ``AlexNet'' network.\footnote{
  This can be downloaded using the Caffe tools from the ``model zoo.''
  \url{http://caffe.berkeleyvision.org/model_zoo.html}
}
The network was progressively sparsified in stages over 200,000 iterations,
taking approximately one week on a GeForce GTX TITAN.
Thresholding/$\ell_0$ projection was done every 100 iterations.
The number of nonzeros remaining after the thresholding was manually reduced
in steps, where we tightened the sparsity constraint each time the
$\ell_0$-constrained training converged on a network with comparable accuracy
to the original dense weights.

\subsection{Optimization}

As a first experiment, we verify that the stochastic gradient descent solver
is able to successfully optimize the problems given by our new regularizations.
This can be seen by looking at the training loss and accuracy,
plotted in Figure \ref{fig:optim}.
In this optimization, we considered training a baseline CIFAR-10 model
with sparsity penalties on the fully-connected layers.

Optimization with the $\ell_0$-norm projection was seen to be far more robust
than the authors' expectations.
We did not observe any case where a network with baseline regularization
was successfully trained but an $\ell_0$-regularized variant of the model
failed to find a model reasonably close the optimum w.r.t.~the training loss.
We hypothesize that the techniques utilized by existing Deep Learning training
methods to optimize the highly nonconvex objectives of those
models are already powerful enough to in practice cope with the
additional complexity incurred by the nonconvex feasible region.

This held across all choices of layers, structured updates, and other
hyperparameters.  In addition to this experiment,
the surprisingly high test accuracies for very sparse models seen in this
work's other experiments also suggest that the optimization
was successful in finding a reasonable fit to the training data in those cases.

\input{fig/optim/optim.tex}

\subsection{Layer-wise distribution of sparsity}
\label{sec:greedy-layerwise}

We observed on standard experimental networks that each layer
of the network performs very differently as we vary the level of
sparsity-inducing regularization.
It can for instance be seen in Figure \ref{fig:partial-sparsity} that
significant decreases in accuracy are seen when sparsifying the first
convolutional layer at nonzero ratios for which the fully-connected
layers are still able to describe the target
concept.
The distribution of sparsity between different layers is the hyperparameter
with the greatest effect on the performance of sparse deep models.
The procedure below allows us to automatically determine sparsity
hyperparameters incorporating some of the intuitions given by the
layer-specific results in Figure \ref{fig:partial-sparsity}.

The choice of which layers to sparsify was done by doing a greedy search
on the parameter space, trying to reduce the number of nonzeros while
maximizing the accuracy on a validation set.
The procedure was:
\begin{enumerate}
  \item Separate a validation set from the training data.
  \item Repeat until a model of the desired sparsity is found:
  \begin{enumerate}
    \item For each layer, reduce the number of nonzeros by 20\% and train a network.
    \item Remove 20\% nonzeros from the layer whose network in (a) produce the best validation accuracy.
    \item Use this network as the start of the next iteration.
  \end{enumerate}
\end{enumerate}

We noted that for a baseline CIFAR-10 network
this procedure yielded nonzero distributions heavily skewed
toward sparsifying the later layers of the network.
The level of sparsity, and the number of nonzero parameters, chosen by
this procedure for each layer of the best candidate networks is shown
in Figure \ref{fig:nnz-distribution}.
From the normalized plot on the right of Figure \ref{fig:nnz-distribution},
it is clear that the tightest sparsity constraints were imposed on the
final convolution layer and the fully-connected layers
while the first two convolution layers were relatively untouched.
In terms of the number of parameters set to zero,
the greatest reduction in weights was in the final convolution and first
fully-connected layers.
We found that these networks significantly outperformed baselines that
imposed sparsity constraints uniformly across all layers.

\input{fig/nnz-distribution/nnz-distribution.tex}

%% file: fig/optim/optim.tex
\begin{figure}
  \includegraphics[width=0.5\linewidth,clip=true,trim=0 0 0 0]{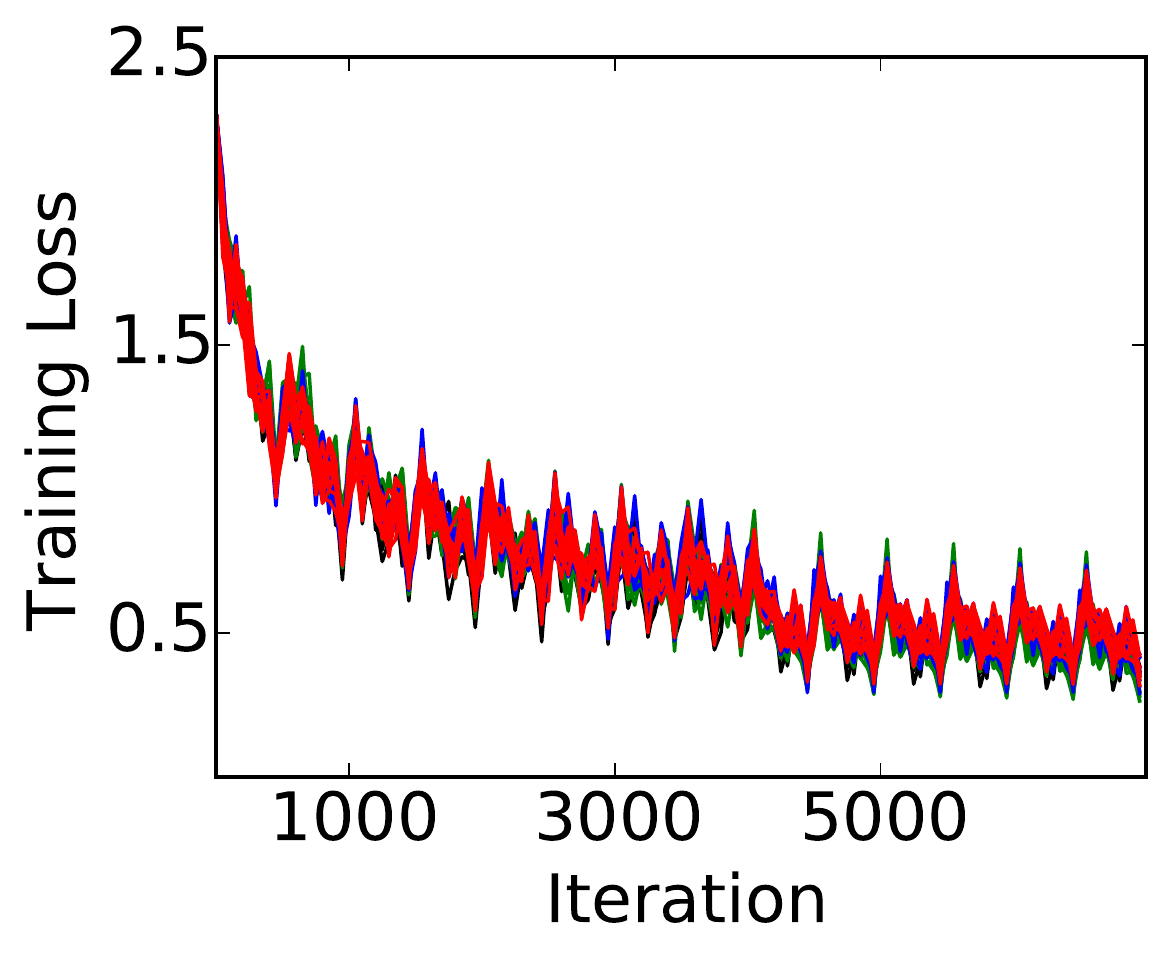}
  \nolinebreak[4]
  \includegraphics[width=0.5\linewidth,clip=true,trim=0 0 0 0]{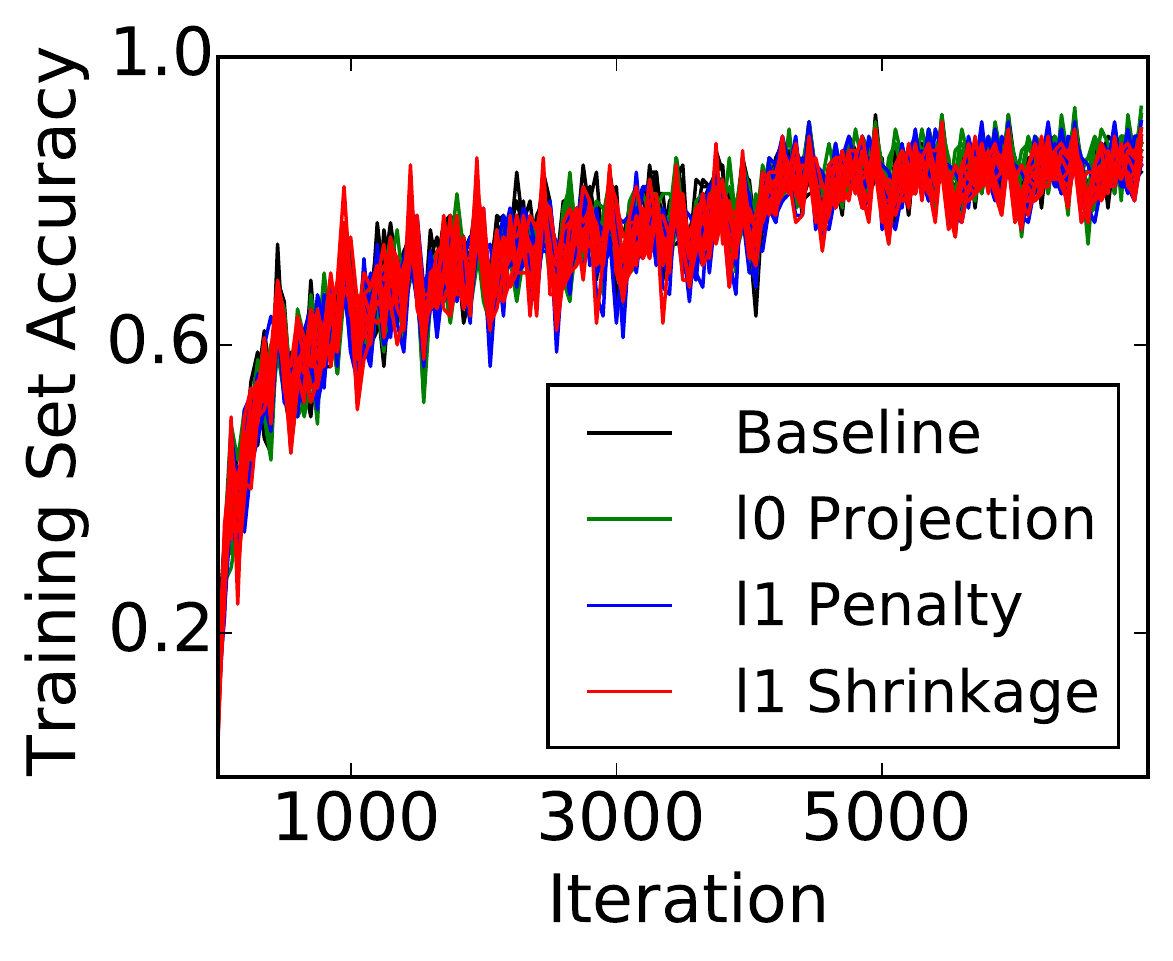}
  
  \caption{ \label{fig:optim}
    This figure plots the progress of the stochastic gradient optimization. 
    We plot 10 repetitions of each method, we found that the optimization
    consistently converged when using any of the sparsity updates presented
    in this work.
  }
\end{figure}

%% file: fig/nnz-distribution/nnz-distribution.tex
\begin{figure}
  \begin{center}
    \includegraphics[height=0.38\linewidth]{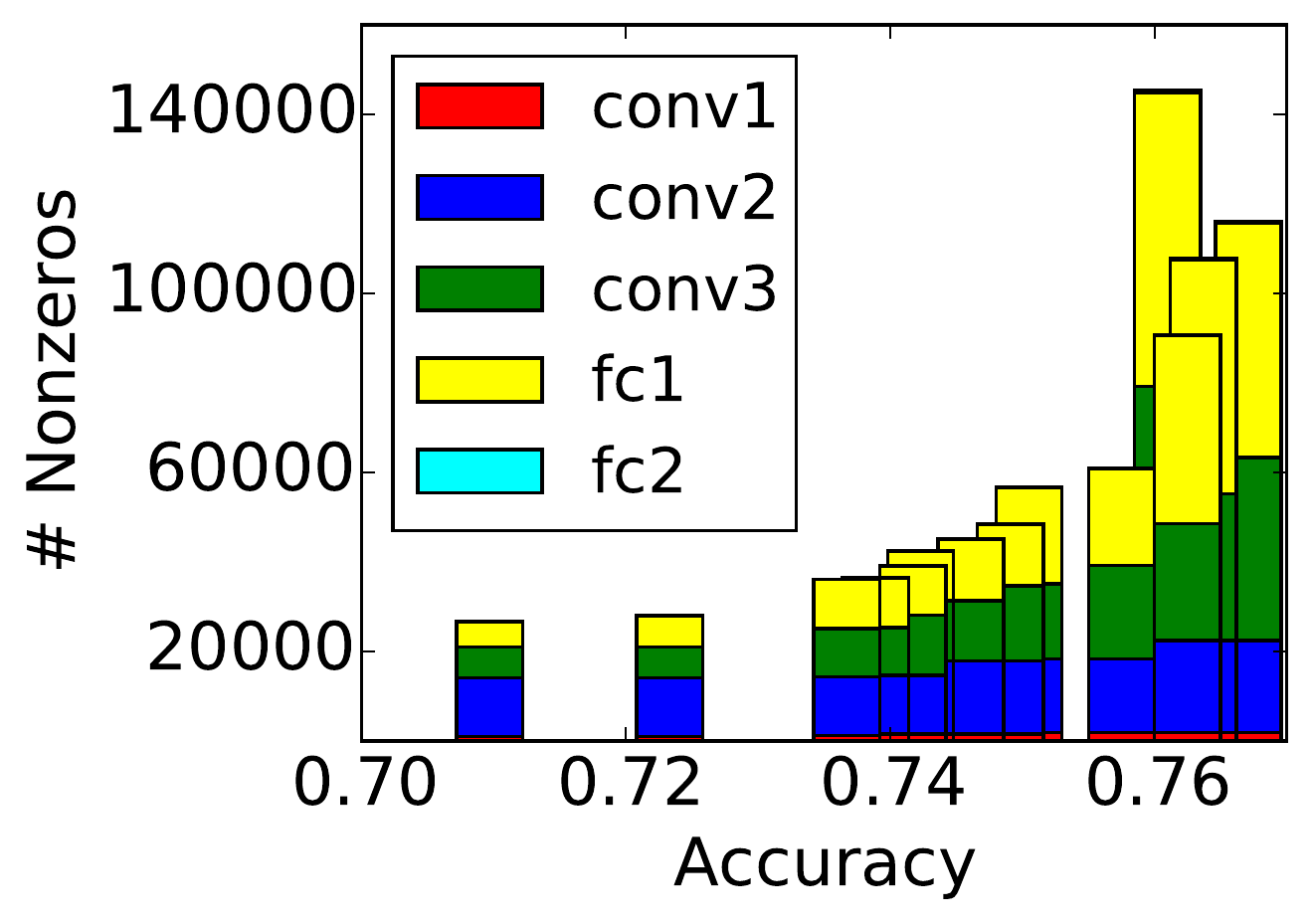}
    \nolinebreak[4]
    \includegraphics[height=0.38\linewidth]{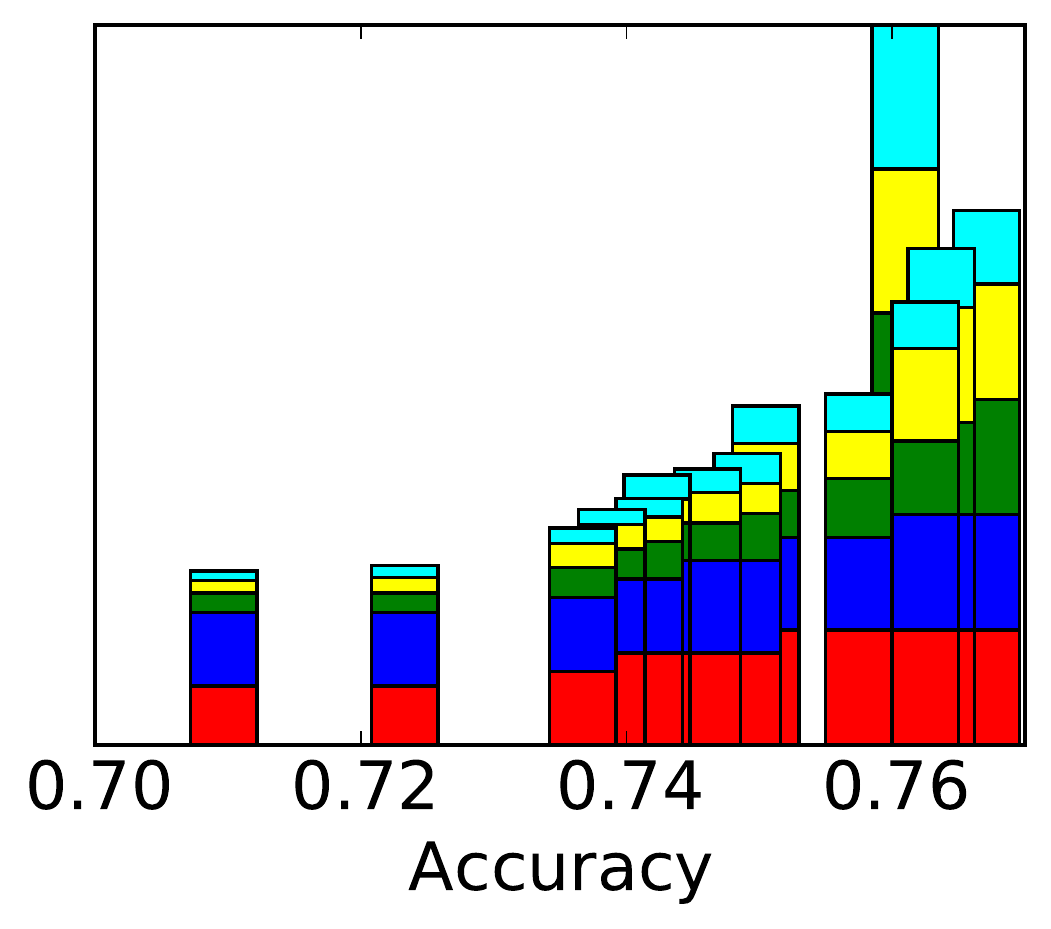}
  \end{center}

  \caption{ \label{fig:nnz-distribution}
    These plots show the distribution of nonzero parameters determined
    by a greedy procedure seeking to maximize the accuracy of sparse networks
    for CIFAR-10.
    Each stack of boxes corresponds to a single network,
    and is centered on the accuracy for that network.
    The plot on the left directly counts the number of nonzeros in the layer.
    The plot on the right shows the same networks, but
    normalizes the height of the boxes such that
    each layer's box would be the same height for a dense network.
    ``conv1-3'' are the convolution layers,
    while ``fc1-2'' are the fully-connected layers.
  }
\end{figure}

%% file: experiments.tex
\section{Experiments}
\label{sec:experiments}

We used MNIST \cite{lecun1998-document-recognition},
CIFAR-10 \cite{krizhevsky2009-tiny-images},
and ImageNet \cite{deng2009-imagenet}
as benchmarks,
and show the effect of sparsity on widely-distributed baseline CNNs
for these datasets.

MNIST is a set of handwritten digits, with 60,000 training examples and
10,000 test examples.
Each image is $28 \times 28$ with a single channel.
MNIST is among the most commonly used datasets in machine learning,
computer vision, and deep networks.

CIFAR-10 is a subset of the ``80 million tiny images'' dataset
\cite{torralba2008-tinyimages},
where ground truth class labels have been provided.
There are 10 classes, with 5,000 training images and 1,000 test images per
class, for a total of 50,000 training images and 10,000 test images.
Each image is RGB, with $32 \times 32$ pixels.
Other than subtracting the mean of the training set,
we do not consider whitening
or data augmentation, obtaining higher absolute error rates but allowing us to
compare against an easily reproducible baseline.

The AlexNet fine-tuning experiments used the ILSVRC 2012 training set.
Simple data augmenation using random crops and horizontal mirroring
was used in the training phase, along with subtracting the mean image
from the training set.

\subsection{Thresholding}

Convolutional Neural Nets trained with existing $\ell_2$ regularization methods
based on weight decay will have many weights of very small
magnitude, due at least in part to the number of redundant parameters
that never receive large updates during back-propagation.
These networks do not however approximate a sparse network,
nor can they be made sparse without additional training using regularizations
like those presented in this work.
In Figure \ref{fig:thresholding} we present the result of an experiment
comparing models trained using our $\ell_0$ projection to a model using
ordinary weight decay.
In this experiment, we first threshold weights with magnitude less than
$\delta$, for varying choices of $\delta$ in the interval $[0,0.1]$,
on an $\ell_2$-regularized model trained on CIFAR-10.
The distribution of nonzero parameters between layers in the thresholded
model is then duplicated in a model trained with $\ell_0$ regularization.
This is done by imposing a per-layer $\ell_0$ constraint through our projection
operator, where the maximum number of nonzeros allows is the same as in
that layer of the thresholded model.
This resulting test accuracies show that a far more useful model may be
obtained at high levels of sparsity if the thresholding is alternated
with training, as in our $\ell_0$ projection method,
rather than done only as a post-processing step.


\input{fig/thresholding.tex}

\subsection{Accuracy and Regularization Updates}

\input{fig/partial_sparsity.tex}

\paragraph{MNIST and CIFAR-10}
A key empirical result of this work is that sparse models achieve surprisingly
high accuracies even as the number of nonzero parameters get quite small.
In Figure \ref{fig:partial-sparsity} we look at
limited experiments where particular layers of two baseline MNIST and CIFAR-10
models are sparsified with various penalties.
We consider as a baseline both thresholding $\ell_2$-regularized models
and varying the network structure to directly reduce the number of parameters.

We note that, while the $\ell_0$ projection is the regularization that
imposes sparsity most directly,
the $\ell_1$-based regularizations perform comparably well for a ``middle''
range of regularization strength.
This is as would be expected, based on the literature for sparsity-inducing
regularization for shallow models.
The $\ell_0$ projection tends to outperform the $\ell_1$ subgradient
and shrinkage updates as the number of nonzeros begins to approach within
an order of magnitude of the dense model.
This can be explained as the result of the $\ell_1$-based updates' effect
on the \emph{magnitude} of the regularized weights,
by contrast to the $\ell_0$ projection that does not at all modify the
largest-magnitude weights.
This distinction becomes more important as you have more smaller-magnitude weights in less sparse models.
It is also less visible in MNIST as all models quickly reached a ``ceiling''
accuracy.
In the other direction, the hard thresholding/$\ell_0$ update seems to produce
models of moderate accuracy at higher levels of sparsity.
In this range, for higher regularization strength, the $\ell_1$ updates also
frequently fail to produce very sparse models at all,
due to a discontinuity in the regularization path between the sparser
models shown in \ref{fig:partial-sparsity} and models that are all zero.

Our later experiments primarily use the $\ell_0$ projection because it
requires far less intensive hyperparameter tuning.
The different sparsity levels seen for the $\ell_1$ norms are only
seen in a very narrow range of values for the regularization multiplier.
Outside this range we see either dense models or models that are
completely zero.
By contrast, good models are produced for nearly any choice of the number of
nonzeros imposed by an $\ell_0$ constraint.

\paragraph{ImageNet}
In the AlexNet experiments
we focus on the case of applying our $\ell_0$-projection regularization,
and only on the fully-connected layers as these are together responsible
for 96\% of the total number of weights in the network.

The Top-1 and Top-5 validation accuracies of the original networks,
the Caffe duplication, and our sparse version of the Caffe duplication
are shown in the table in Figure \ref{fig:memory}.
Note that the Caffe duplication achieves slightly lower validation accuracies
than the original networks trained by Krizhevsky~\textit{et.~al.}.
Initially, only the final fully-connected layer was sparsified and reduced
to 400,000 nonzero weights out of 4,096,000 in the original dense network
(plus 1,000 per-neuron biases).
This is \emph{ten times smaller} than the original parameter set of this layer
in the dense network.
In the second stage, the other two fully-connected layers were also
regularized to produce a network with 3 million parameters in each of these two
layers (for 6 million total).
This is from a total of 54.5 million weights in these two layers of the
original dense network.
Overall, our sparse network has all but 14\% of the weights of the network
set to zero.

To compare, we also directly thresholded the network without additional
training to have the same number of nonzeros in each layer.
This yields a network that achieves only 38.92\% Top-1 validation
accuracy and 63.44\% Top-5 validation accuracy.
The accuracies reported here are done without test-time oversampling
or any similar test-time data augmentation methods.

\subsection{Ensembles}

Our results suggest that the very high parameter counts in deep learning
models include a number of redundant weights.
Much of the computational resources and model complexity incurred by these
very large models is spent to yield relatively little benefit in terms
of test-time accuracy.
Given a fixed budget of some computational resource (e.g.~memory),
this is much better spent on more effective ways to increase accuracy.
A particularly powerful method for improving test-time accuracy,
known to work quite well when maximizing accuracy on difficult machine
learning benchmarks \cite{bell2007-netflix}
is building \emph{ensembles} that combine the output of multiple models.

\input{fig/ensembles.tex}
\input{fig/data-starvation/data-starvation.tex}

We construct ensembles using \emph{bagging} \cite{breiman1996-bagging},
a classical Machine Learning technique in which each member of the ensemble
is trained on a random resampling of the training data.
Bagging is frequently used where the ensemble members are
expected to overfit or produce unstable predictions.
In Table \ref{tab:ensembles} we show the resulting accuracy as we grow
the ensemble under a \emph{parameter budget}.
For each ensemble we train, we seek to maintain a set of nonzero weights
that does not grow in size even as we consider greater numbers of predictors
in the ensemble.
This can be done by sparsifying the individual elements of the ensemble
with the regularizations presented in this work.
This targets e.g.~a setting such as a mobile device with limited memory,
for which building ensembles with the full model is prohibitively expensive;
which can be true quite quickly for CNNs.
Using this as a proxy for model capacity and power, this experiment further
allows us to explore the tradeoffs in how we build predictors and where
model capacity should be spent to get the best performance at deployment.
To train an ensemble with $n$ members, we repeat the following steps
$n$ times:
\begin{enumerate}
  \item Resample the training data, with replacement, to get another training set of the same size.
  \item Take a layer-wise distribution of nonzeros given by the method in Section \ref{sec:greedy-layerwise} that has $\le 1/n$ total nonzeros.
  \item Train a model on the resampled data, with each layer having an $\ell_0$ constraint to enforce this distribution.
\end{enumerate}
At test time, we run each CNN as normal over the test dataset.
We then average the output layers, and predict the class
corresponding to the largest average output.

We observe a significant increase in accuracy for smaller ensembles.
As the size of the ensemble grows and the models become sparser, however,
this yields diminishing returns as the individual models can no longer
sufficiently approximate the target task.


\subsection{Reduced Training Data}
\label{sec:data-starvation}

A key property of regularization is its ability to improve the
\emph{generalizability} of a learned model.
If there is insufficient training data to properly estimate the true
underlying distribution,
the minimizer of the empirical risk will be different from the
best model for the expected risk under the true distribution.
The model will then ``overfit'' the training data rather than correctly
learning the target concept.

In Figure \ref{fig:data-starvation}, we test this assertion using models
trained with our regularization and with $\ell_2$ weight decay.
We randomly subsample the CIFAR-10 training dataset, to produce a series
of smaller datasets with increasingly insufficient training data.
Two results can be seen from this experiment that match what Machine Learning
theory predicts.
First, the differences between simpler and more complex models become narrower
with less training data, on the left-hand side of the plots.
Indeed, the simpler models begin to outperform the more parameter-heavy
dense models in some cases when little training data is available.
Secondly, in all models we see as more training data becomes available that
the training accuracy decreases while the test accuracy on unseen data
increases.
The change in accuracy on both sets is much less pronounced on simpler models.

%% file: fig/thresholding.tex
\begin{figure}
  \begin{center}
    \includegraphics[width=1\linewidth,clip=true,trim=25 10 32 10]{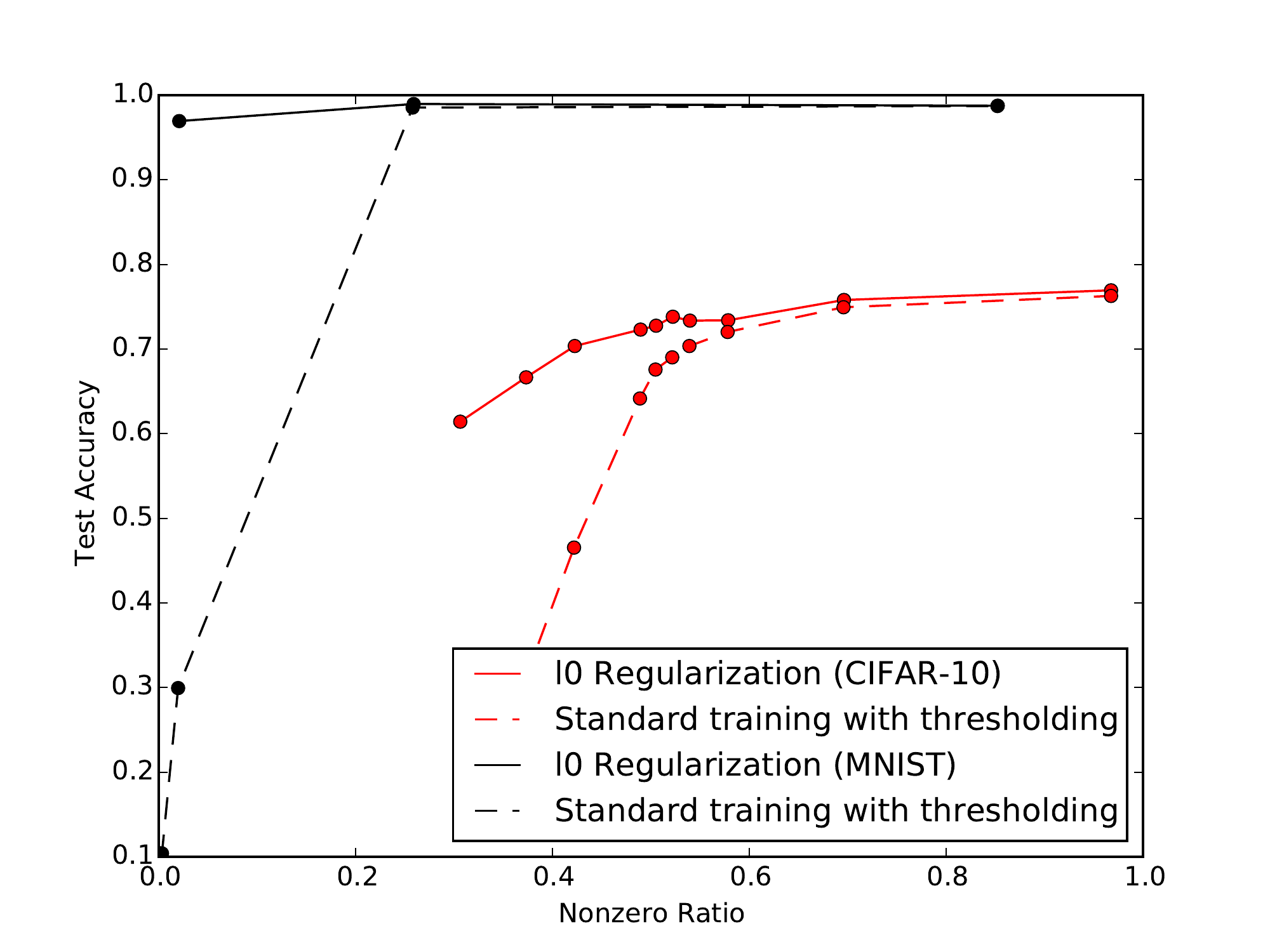}
  \end{center}

  \caption{ \label{fig:thresholding}
    While the individual updates used in the optimization under
    $\ell_0$-projection are a sort of thresholding operator,
    it behaves very differently from simply thresholding the model.
    We see that allowing the model to optimize under an $\ell_0$
    constraint improves the end test accuracy as compared
    to simple thresholding.
  }
\end{figure}

%% file: fig/partial_sparsity.tex
\begin{figure*}
  \begin{center}
    \begin{tabular}{c c c c c}
      \multicolumn{2}{c}{MNIST}
      & & \multicolumn{2}{c}{CIFAR-10} \\
      \cline{1-2} \cline{4-5}
      \hspace{0.9em} $1^\text{st}$ Convolution Layer & \hspace{1.4em} Fully-connected
      & & \hspace{0.9em} $1^\text{st}$ Convolution Layer & \hspace{1.4em} Fully-connected \\
      \includegraphics[width=0.225\linewidth,clip=true,trim=0 0 0 0]{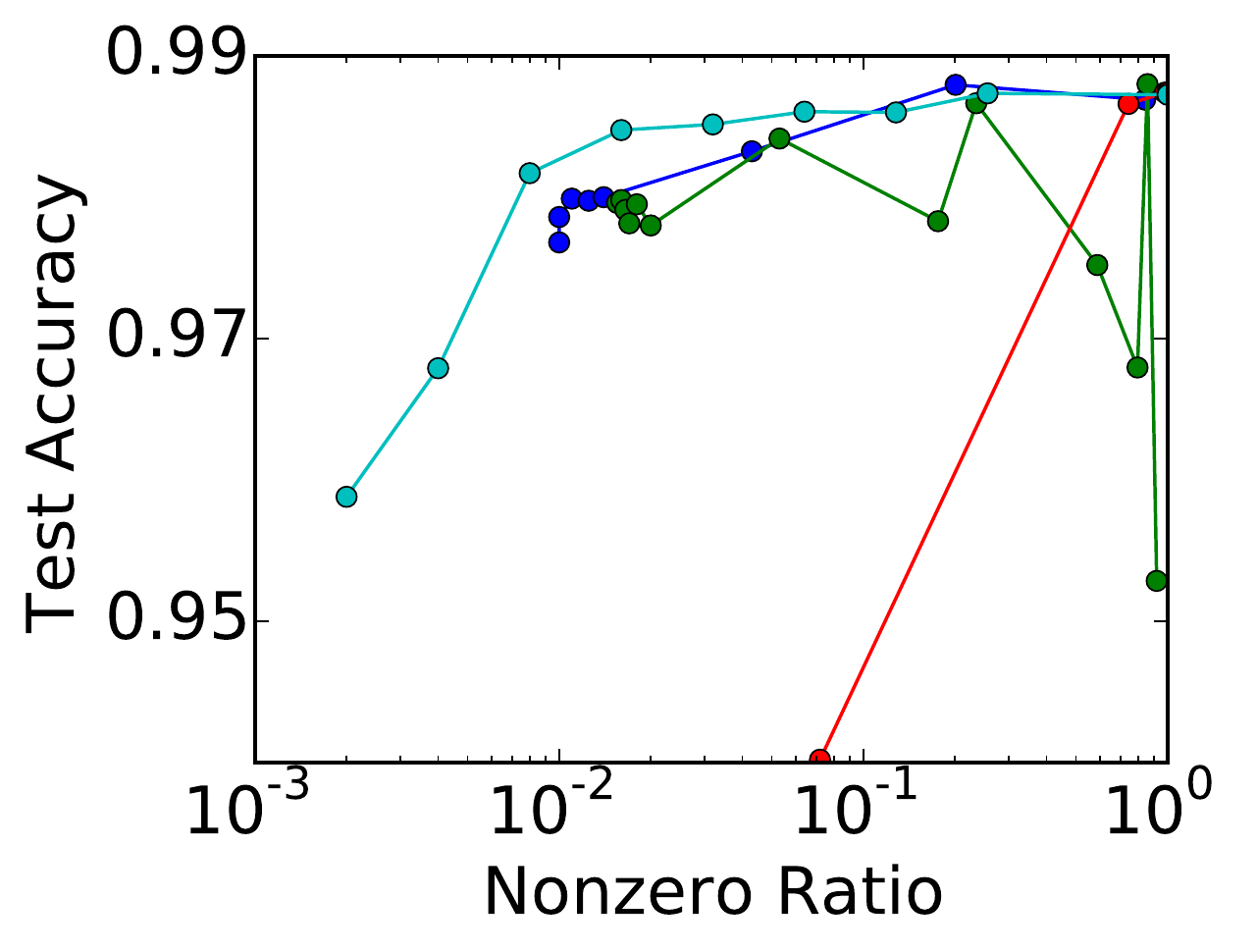}
      & \includegraphics[width=0.225\linewidth,clip=true,trim=0 0 0 0]{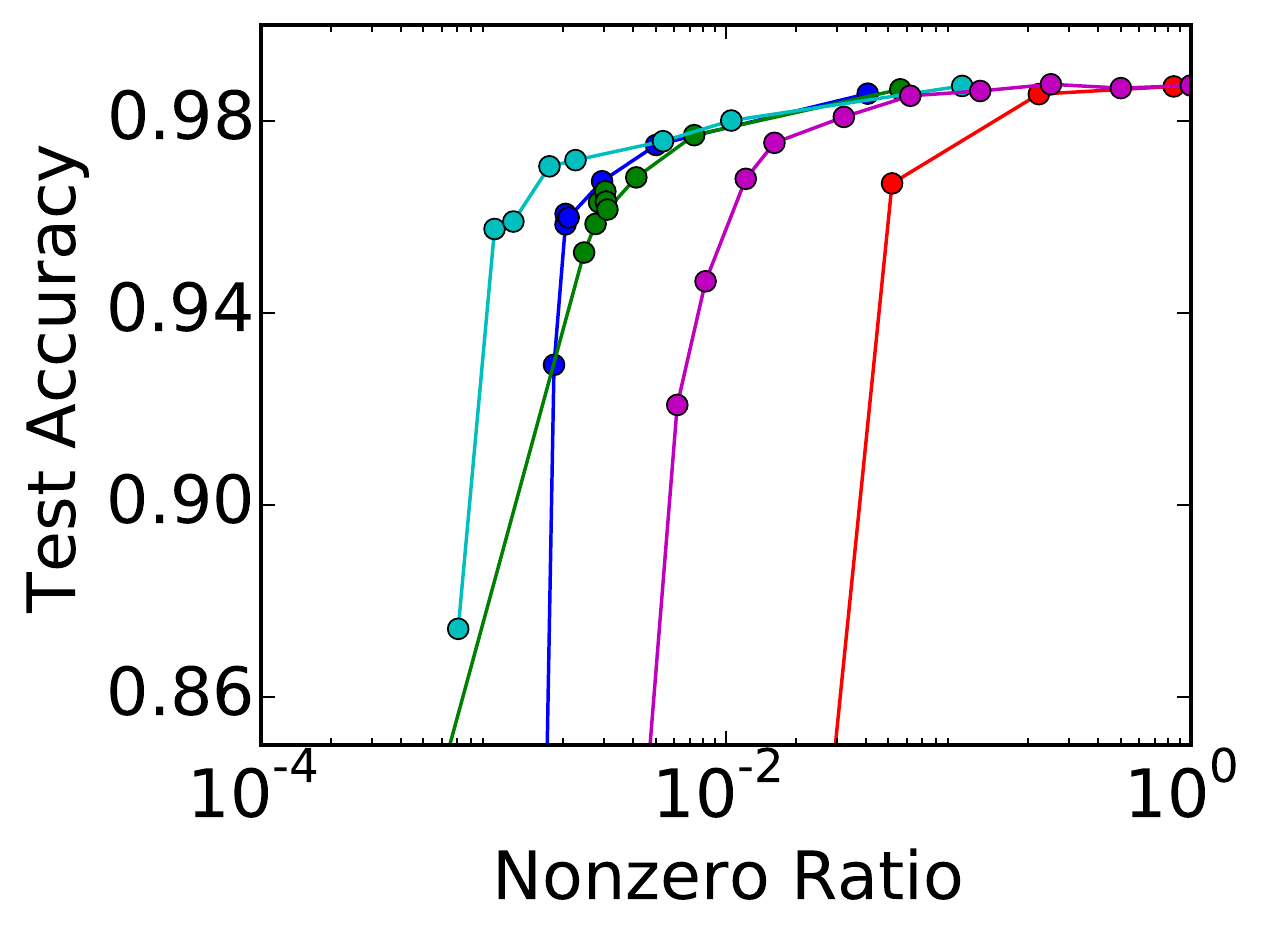}
      & & \includegraphics[width=0.225\linewidth,clip=true,trim=0 0 0 0]{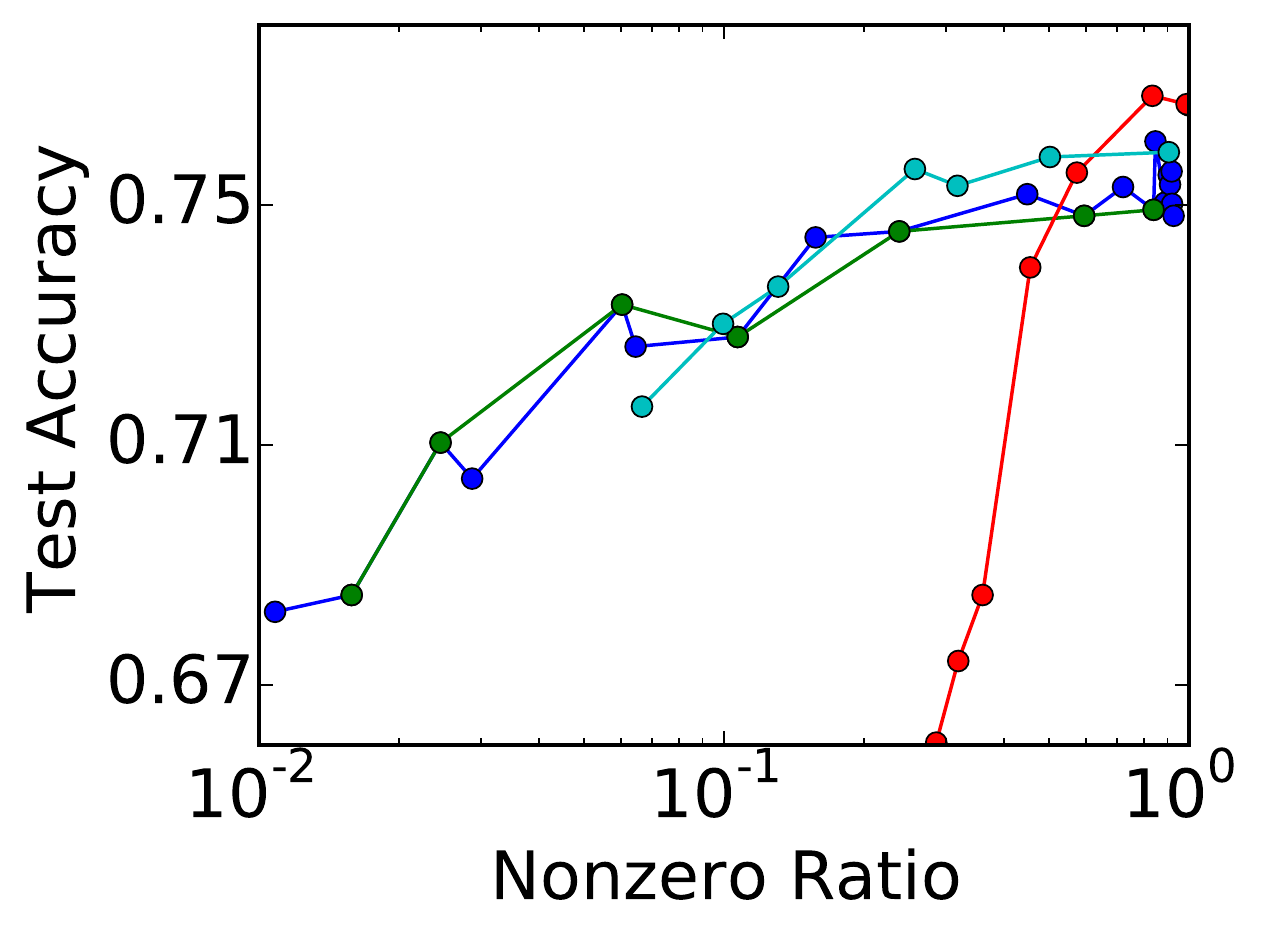}
      & \includegraphics[width=0.225\linewidth,clip=true,trim=0 0 0 0]{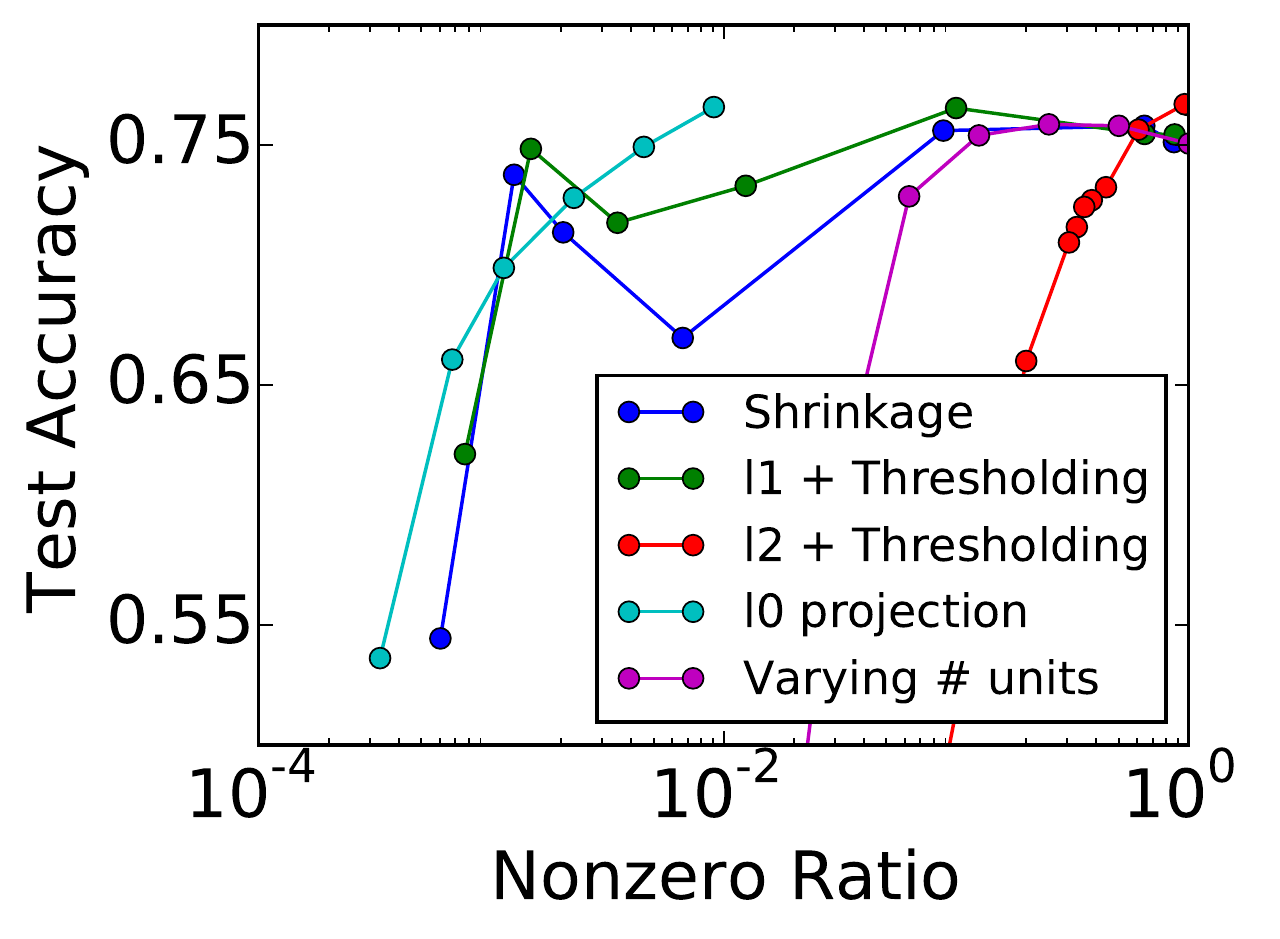}
    \end{tabular}
  \end{center}

  \caption{ \label{fig:partial-sparsity}
    Sparsity vs. accuracy when some of our regularizations are imposed
    only on specific layers of the MNIST and CIFAR-10 baseline networks.
    As baselines we also consider both thresholding the weights learned
    under $\ell_2$-regularization,
    and reducing the number of
    hidden units in the first fully-connected layer.
  }
\end{figure*}

%% file: fig/ensembles.tex
\begin{table}
  \begin{tabular}{l r r r}
    \textbf{Nets} & \textbf{Nonzeros/Net} & \textbf{Total} & \textbf{Test Accuracy} \\
    \hline
    1 & 145578 & 145578 & 75.85\% $\pm$ 0.559\% \\
    2 & 71770 (49.3\%) & 143540 & 77.40\% $\pm$ 0.192\% \\
    3 & 46023 (31.6\%) & 138069 & 77.18\% $\pm$ 0.215\% \\ 
    4 & 36333 (25.0\%) & 145332 & 75.96\% $\pm$ 0.116\% \\
    5 & 28249 (19.4\%) & 141245 & 74.60\% $\pm$ 0.155\% \\
    \hline
  \end{tabular}

  \vspace{2mm}
  
  \caption{ \label{tab:ensembles}
    Building ensembles of sparse models under a parameter budget.
    The 1-model case was trained on the original dataset as a baseline,
    all others on bagged resampling.
    Accuracies given are means with standard deviations across multiple trials
    with different models and bagged datasets in each trial.
  }
\end{table}

%% file: fig/data-starvation/data-starvation.tex
\begin{figure}
  \begin{center}
    \includegraphics[width=0.49\linewidth]{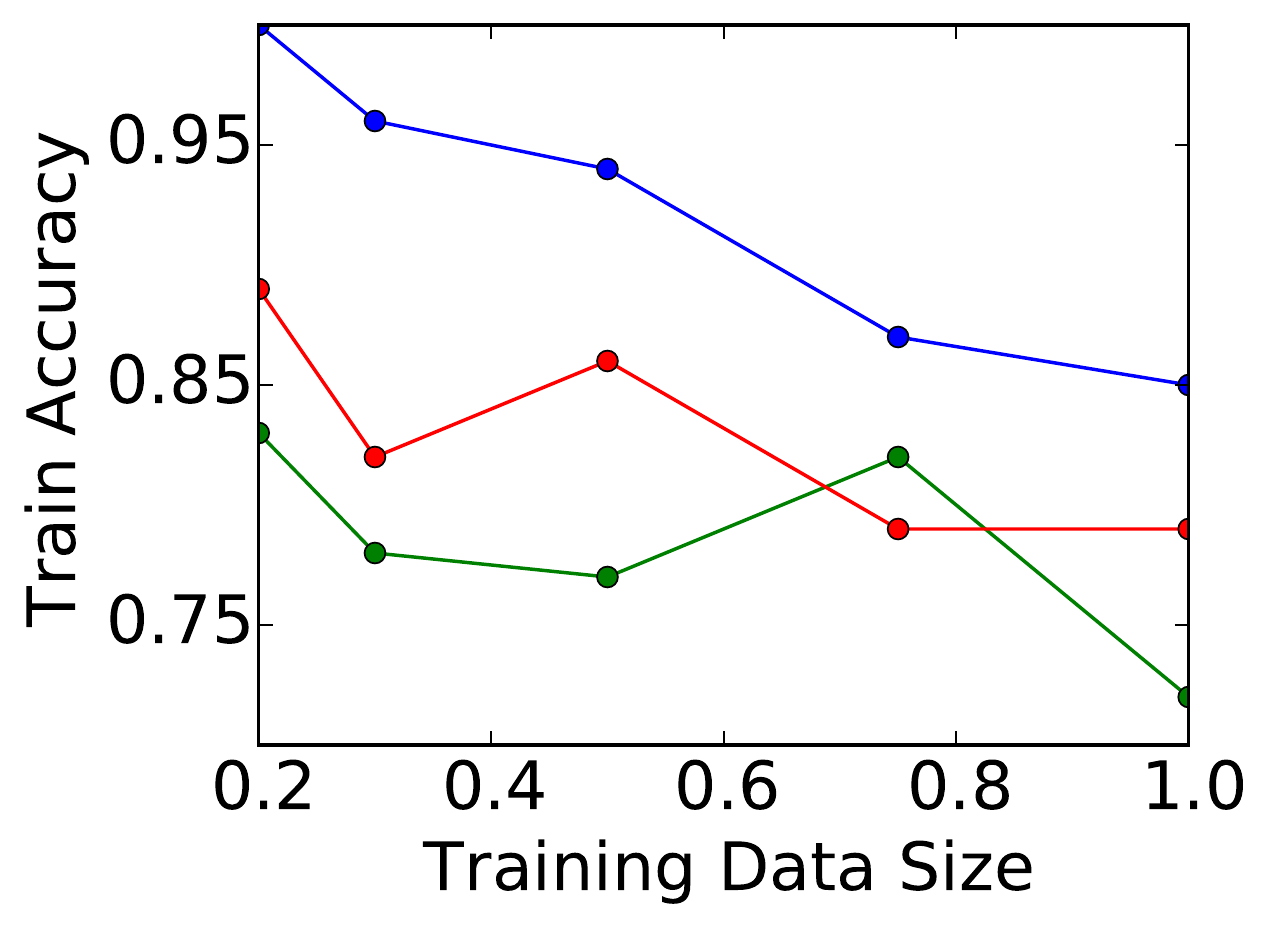}
    \nolinebreak[4]
    \includegraphics[width=0.49\linewidth]{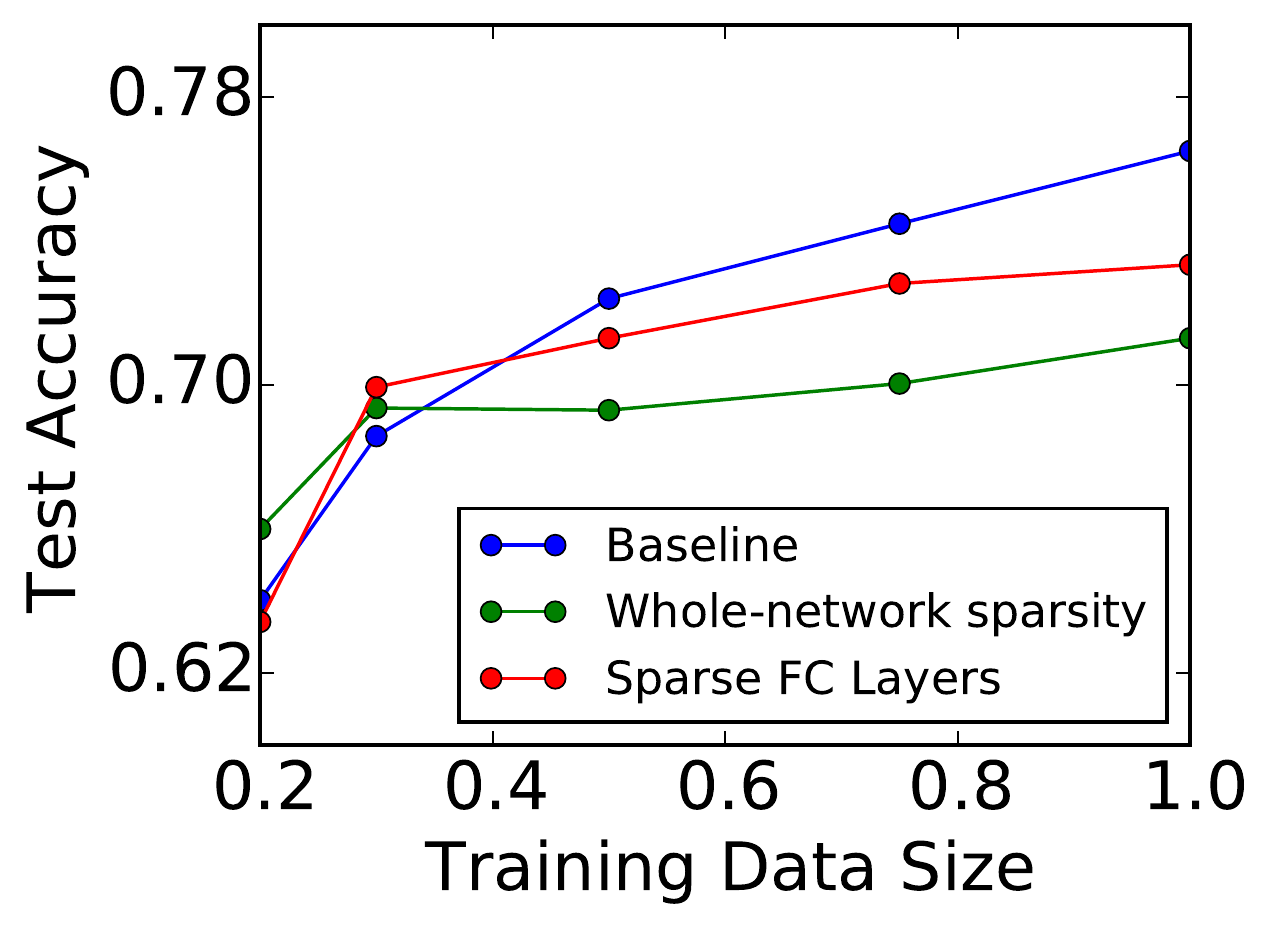}
  \end{center}
  
  \caption{ \label{fig:data-starvation}
    Training and test accuracy for dense and sparsity-regularized
    CNNs as we vary size of the training set by randomly subsampling CIFAR-10
    as in Section \ref{sec:data-starvation}.
  }
\end{figure}

%% file: discussion.tex
\section{Discussion}

In this work we present a powerful technique for regularizing and constructing
simpler deep models for vision.
This may be a key tool in training deep learning vision models that are meant
to be deployed in resource-constrained environments,
or to increase the generalizability of existing models.
Using sparsity-inducing regularization provides a significantly improved
method for reducing the parameters of a model as compared with baselines
including reducing the number of units in the network and simple thresholding.

An interesting empirical observation we see is that very sparse models
manage to do surprisingly well on benchmark image classification tasks.
We leverage this not only to construct very simple models,
but also to build ensembles of
these sparse models that outperform the baseline dense models while
still staying within fixed resources.

%% file: supp-memory.tex
\section{Memory Usage}
\label{sec:memory}

To describe what the sparsity levels mean in concrete computation terms
we consider three storage formats, each of which allow for fast computation
with sparse networks.
When calculating the memory usage of sparse networks,
we assume the optimal choice among the following formats:
\begin{enumerate}
\item \textbf{Dense:}
  One can ignore any sparsity present in the network, and store it as
  with the original dense weights.
  For very sparse networks a great deal of storage will be filled with
  the number zero, but for networks with few zero weights this will still
  be cheaper due to the additional overhead for required sparse data
  structures.
\item \textbf{Bitmask:}
  We have a simple bitmask with a number of bits equal to the number of
  total parameters.
  Iff the bit corresponding to a a parameter is one.
  then there the value of that parameter is nonzero.
  The nonzero parameters are then stored in a flat array.
  Additional indices into the flat array, depending on how computation
  is done with these weights, can allow this format to still be used
  directly in CNNs at runtime.
\item \textbf{Indexed:}
  Only the nonzero parameters are stored, each one as a pair consisting
  of an index into the original weight array alongside the weight value.
  This is the traditional way to handle sparse vectors.
\end{enumerate}
In some types of layers more specific sparse formats such as
Compressed Sparse Row (CSR) matrices may also be suitable for computation.
These will have a memory cost approximately equal to the ``indexed'' case.


We assume that the weights are stored as single-precision (32-bit/4-byte)
floating-point values.
The overhead introduced by the sparse data structures is relatively smaller
for double-precision (64-bit/8-bytes) floating-point values.
Additional formats, such as half-precision floats
or fixed-point weights are not supported by
standard hardware or CNN codes.

Using these formats, we plot in Figure \ref{fig:memory} the memory required
to store the weights of sparsified forms of the baseline test networks.
Each point is a candidate network considered
by a greedy search over the per-layer distribution of nonzeros.
In the table in the same figure, we give the memory used for the ``indexed''
format on a sparse CNN for ImageNet.
We use Kibibytes ($2^{10}$ bytes) and Mebibytes ($2^{20}$ bytes),
abbrevated to ``KB'' and ``MB'' respectively.